%% file: 0.main.tex
\documentclass[conference]{IEEEtran}
\IEEEoverridecommandlockouts
\usepackage{cite}
\usepackage{amsmath,amssymb,amsfonts}
\usepackage{algorithmic}
\usepackage{graphicx}
\usepackage{textcomp}
\usepackage{xcolor}
\usepackage{url}
\usepackage{amssymb}
\usepackage{bm}
\usepackage{times}
\usepackage{algorithmic}
\usepackage{amsmath}
\usepackage{booktabs}
\usepackage{multirow}
\usepackage{subfigure}
\usepackage{graphicx}
\usepackage{algorithmic}
\usepackage[linesnumbered, ruled, vlined]{algorithm2e}
\usepackage{bm}
\usepackage{mathrsfs}
\usepackage{multirow}
\let\labelindent\relax
\usepackage{enumitem}
\usepackage{threeparttable}
\usepackage{footnote}
\usepackage{times,flushend}
\usepackage[square, comma, sort&compress, numbers]{natbib}
\usepackage[normalem]{ulem} 

\input{math_commands.tex}

\def \revise#1{{\textcolor{black}{#1}}}
\begin{document}\small

\title{Inductive Granger Causal Modeling for Multivariate\\ Time Series}

\author{
\IEEEauthorblockN{
Yunfei Chu,
Xiaowei Wang,
Jianxin Ma,
Kunyang Jia,
Jingren Zhou
and
Hongxia Yang \\
Alibaba Group, China}
\IEEEauthorblockA{
Email: \{fay.cyf, daemon.wxw, jason.mjx, kunyang.jky, jingren.zhou, yang.yhx\}@alibaba-inc.com}}
\maketitle

\begin{abstract}
Granger causal modeling is an emerging topic that can uncover Granger causal relationship behind multivariate time series data. In many real-world systems, it is common to encounter a large amount of multivariate time series data collected from different individuals with sharing commonalities. However, there are ongoing concerns regarding Granger causality's applicability in such large scale complex scenarios, presenting both challenges and opportunities for Granger causal structure reconstruction. Existing methods usually train a distinct model for each individual, suffering from inefficiency and over-fitting issues. To bridge this gap, we propose an \textbf{In}ductive \textbf{GR}anger c\textbf{A}usal modeling (\textbf{InGRA}) framework for inductive Granger causality learning and common causal structure detection on multivariate time series, \revise{which exploits the shared commonalities underlying the different individuals. In particular, we train one global model for individuals with different Granger causal structures through a novel attention mechanism, called prototypical Granger causal attention. The model can detect common causal structures for different individuals and infer Granger causal structures for newly arrived individuals.} Extensive experiments, as well as an online A/B test on an E-commercial advertising platform, demonstrate the superior performances of InGRA.
\end{abstract}


\section{Introduction}
\input{1.intro.tex} 
\vspace{-2pt}
\section{Related Work}
\input{2.related.tex}

\vspace{-2pt}
\section{The Proposed Method}
\input{3.method.tex}
\vspace{-3pt}
\section{Experiments}
\input{4.experiment.tex}
\section{Conclusion}
\input{5.conclusion.tex}

\bibliographystyle{plain}

{\footnotesize
\bibliography{causal}}
\end{document}

%% file: math_commands.tex
\usepackage{amsmath,amsfonts,bm}

\def\eqref#1{equation~\ref{#1}}

\def\1{\bm{1}}

\def\va{{\bm{a}}}

\def\vd{{\bm{d}}}
\def\ve{{\bm{e}}}

\def\vg{{\bm{g}}}
\def\vh{{\bm{h}}}

\def\vp{{\bm{p}}}
\def\vq{{\bm{q}}}
\def\vr{{\bm{r}}}

\def\vx{{\bm{x}}}
\def\vy{{\bm{y}}}

\def\mX{{\bm{X}}}

\DeclareMathAlphabet{\mathsfit}{\encodingdefault}{\sfdefault}{m}{sl}
\SetMathAlphabet{\mathsfit}{bold}{\encodingdefault}{\sfdefault}{bx}{n}



%% file: 1.intro.tex
Broadly, machine learning tasks are either predictive or descriptive in nature, often addressed by black-box methods \cite{guo2018survey}. With the power of uncovering relationship behind the data and providing explanatory analyses, causality inference has drawn increasing attention in many fields, e.g. marketing, economics, and neuroscience~\cite{pearl2000causality, peters2017elements}. 
Since the cause generally precedes its effects, known as temporal precedence~\cite{eichler2013causal}, recently, an increasing number of studies have focused on causal discovery from time series data. They are commonly based on the concept of Granger causality~\cite{granger1969,granger1980testing} to investigate the causal relationship with quantification measures. 

In many real-world systems, it is common to encounter a large amount of multivariate time series (MTS) data collected from different individuals. The underlying Granger causal structures of such large scale data often vary~\cite{zhang2017causal,huang2019causal}. For example, in the financial market, the underlying causal drivers of stock prices are often heterogeneous across various sectors. Similar phenomenons have also been observed in the different product sales in E-commerce, e.g. factors that influence the buyers' behaviours usually vary across consumers with different profiles.

To this situation, most existing methods, e.g., VAR, have to train separate and independent models for each individual. When facing massive MTS data from different individuals, they have to train a great many models. Moreover, each model is trained with the data from one individual, suffering from over-fitting, especially for long-tailed ones.  Although some works have been proposed to solve such problems with one model \cite{zhang2017causal,huang2019causal}, they lack the inductive capability to do inference for unseen samples. 

\revise{In practice, we found that there also exists shared causal information among different individuals.}
{For example, one may want to buy several different categories of items for a sports festival at the same time, including clothes, sport accessories and foods.}
Such shared information presents opportunities for causal reconstruction to alleviate the over-fitting problem and to do inductive inference. 
However, it is also challenging to detect common and specific causal structures simultaneously. 

In this paper,  we propose an \textbf{In}ductive \textbf{GR}anger c\textbf{A}usal modeling (\textbf{InGRA}) framework for inductive Granger causality learning and common Granger causal structure detection on multivariate time series data. Our approach builds on the idea of quantifying the contributions of each variable series into the prediction of target variable via a novel designed prototypical Granger causal attention mechanism. In order to ensure that the attention capturing Granger causality, we first design an attention mechanism based on Granger causal attribution of the target series and then perform prototype learning that generates shared prototypes to improve the model's robustness.
Extensive experiments demonstrate the superior causal structure reconstruction and prediction performances of InGRA. In summary, our specific contributions are as follows:
\begin{itemize}
    \item A novel framework that inductively reconstructs Granger causal structures for multivariate time series of multiple individuals.
    \item A prototypical Granger causal attention mechanism that summarizes variable-wise contributions towards prediction and generates prototypes representing common Granger causal structures.
    \item Relative extensive experiments on real-world, benchmark and synthetic datasets as well as an online A/B test on an E-commercial advertising platform that demonstrate the superior performance on the causal discovery comparable to state-of-the-art methods.
\end{itemize}

%% file: 2.related.tex

Recently a considerable amount of work has been proposed for causal inference. Classical methods, such as constraint-based methods~\cite{pearl2000causality,spirtes2000causation,peters2013causal,runge2017detecting,zhang2017causal}, score-based methods~\cite{chickering2002optimal} and functional causal models (FCM) based methods~\cite{shimizu2006linear}, mainly focus on i.i.d data. Under the scope of time series, causal inference is commonly based on the notion of Granger causality~\cite{granger1969,granger1980testing}, and a classical way is to estimate linear Granger causality under the framework of VAR  models~\cite{lutkepohl2005new}. However, existing classical methods fail to uncover causal structures inductively.
Neural network based methods that infer causal relationships or relations that approach causality have gained increasing popularity.
NRI~\cite{kipf2018neural} utilizes a neural relation inference model to infer interactions while simultaneously learning the dynamics. DAG-GNN~\cite{yu2019dag} develops a deep generative model to recover the underlying DAG from complex data.
Attention mechanisms have often been adopted to discover relations between variables. For example, Seq2graph~\cite{dang2018seq2graph}
 discovers dynamic dependencies with multi-level attention.
TCDF~\cite{nauta2019tcdf} studies causal discovery through attention-based neural networks with a causal validation step. 
IMV-LSTM~\cite{guo2019exploring} proposes an interpretable multi-variable LSTM with mixture attention to extract variable importance knowledge.
However, these attention mechanisms provide no incentive to yield accurate attributions~\cite{sundararajan2017axiomatic,schwab2019granger}. 
Thus, we propose a novel attention mechanism based on Granger causal attribution to address the problem.

Since our method utilizes the concept of the prototype to detect common causal structures, another line of related research is about prototype learning. Prototype learning is a form of cased-based reasoning~\cite{slade1991case}, which solves problems for new inputs based on similarity to prototypical cases. Recently prototype learning has been leveraged in interpretable classification~\cite{bien2011prototype,KimRuSh14,snell2017prototypical,LiEtAl18,chen2018looks} and sequence learning~\cite{ming2019interpretable}. We incorporate the concept for Granger causal structure reconstruction on time series data for the first time.


%% file: 3.method.tex
In this section, we formally define the problem, introduce the architecture of InGRA, present the prototypical Granger causal attention with the final objective function.
\subsection{Problem Definition} 
Assuming we have a set of heterogeneous multivariate time series from $N$ individuals, i.e., $\{\mX_i\}_{i=1}^N$, with each consisting of $S$ time series of length $T$, denoted as $\mX_i=(\vx^1_i,\vx^2_i,\dots,\vx^{S}_i)^\mathsf{T}\in\mathbb{R}^{S\times{T}}$, where $\vx_i^s=(x_{i,1}^s,x_{i,2}^s,\dots,x_{i,T}^s)^\mathsf{T}\in\mathbb{R}^{T}$ represents the $s$-th time series of the $i$-th individual. \revise{One of the $S$ series $\vx^{1:S}_i$ is taken as the target series $\vy_i$, and the others are taken as the exogenous series.} We aim to train a model that (1) reconstructs Granger causal structures among variables for each individual; (2) generates $K$ common structures among all the $N$ individuals, each structure represented by \revise{a vector $\vp_k\in\mathbb{R}^S,k=1,...,K$, with each element representing the Granger contribution of each variable towards the target}; and (3) learns a nonlinear mapping to predict the next value of the target variable series for each individual, i.e., $\hat{y}_{i,T+1}=\mathcal{F}(\vx^{1:S}_i)$.

\subsection{Network Architecture}\label{sec:net}
Our InGRA framework consists of two parts: a set of parallel encoders, each predicting the target given the past observations, and an attention mechanism that 
quantifies variable-wise contributions towards prediction. 

\begin{figure}[t]
\centering
\includegraphics[width=0.5\textwidth]{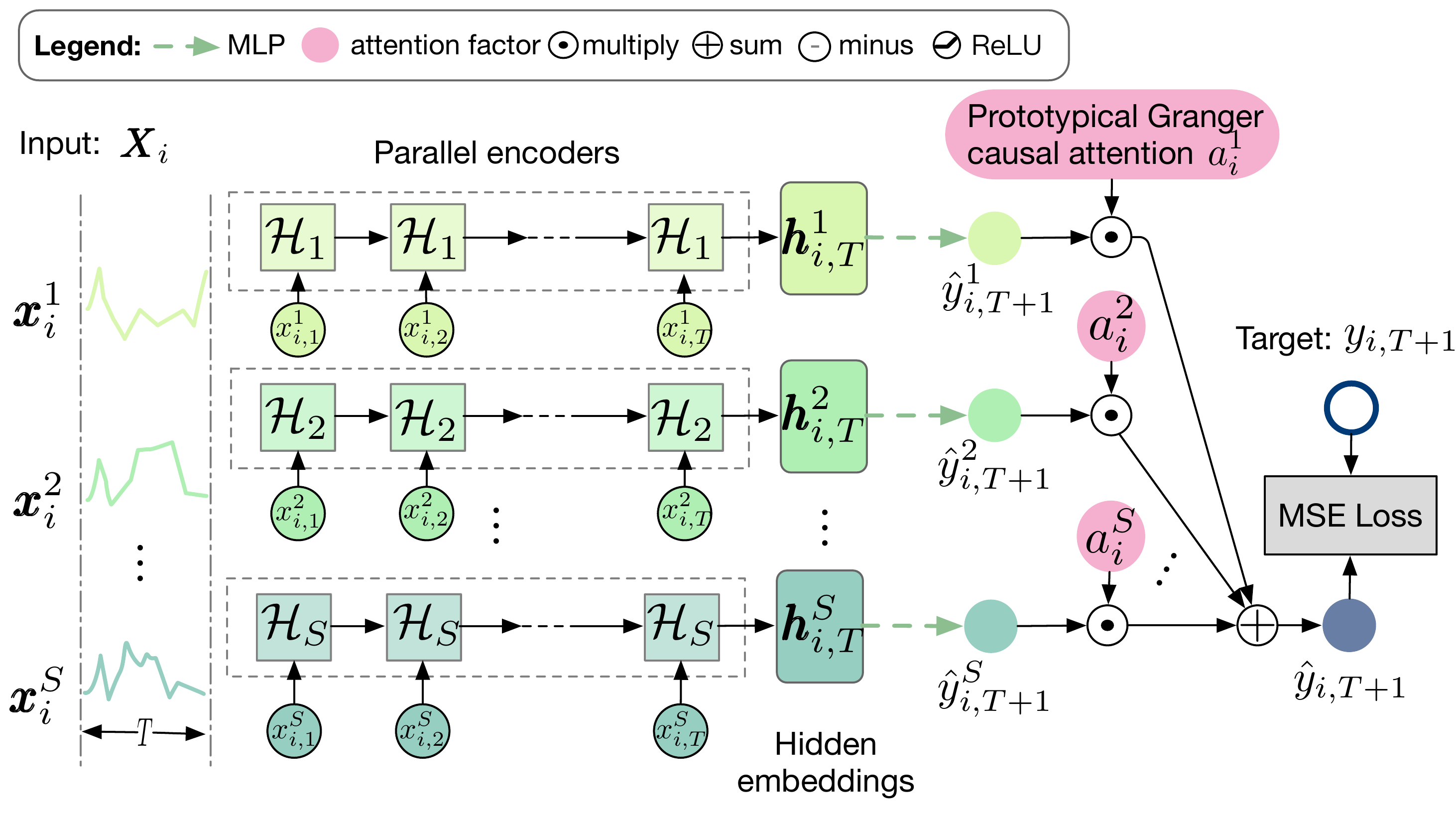}
\caption{The overview of the InGRA framework.}
\label{fig:framework} 
\vspace{-2pt}
\end{figure}

As illustrated in Figure~\ref{fig:framework}, for any input multivariate time series $\mX_i$, the LSTM encoder specific to $s$-th variable projects the time series $\vx_i^s$ into a sequence of hidden state, denoted as $\vh_{i,t}^s=\mathcal{H}_s(x_{i,t}^s,\vh_{i,t-1}^s)$. 
The last hidden states, $\{\vh_{i,T}^s\}_{s=1}^S$, are used as the hidden embeddings. Then the predicted next value of the target variable conditioned on historical data of variable $s$, denoted as $\hat{y}_{i,T+1}^s$, can be computed by $\hat{y}_{i,T+1}^s=f_s(\mathbf{h}_{i,T}^s)$, where $f_s(\cdot)$ denotes the MLP network specific to variable $s$. Then we obtain the prediction $\hat{y}_{i,T+1}$ by aggregating the predicted values specific to variables through the prototypical Granger causal attention described below.

\subsection{Prototypical Granger Causal Attention}
We propose a novel attention mechanism in InGRA, namely prototypical Granger causal attention, to detect Granger causal relationships for each individual and uncover common causal structures among heterogeneous individuals. \revise{The goal is to learn attentions for each individual, so that the attention vectors can reflect the Granger causal strength between the $N-1$ exogenous variables and the target variable, and generate a group of prototypical vectors capturing the shared commonality among different individuals.} The idea of the prototypical Granger causal attention mechanism is as follows. The Granger causal attribution corresponding to each individual is first computed according to the concept of Granger causality (Figure~\ref{fig:gc}), followed by the prototype learning that summarizes common causal structures for heterogeneous individuals in the training set, and produces the prototypical attention vector (Figure~\ref{fig:proto}). The details of these two parts are described below.
\subsubsection{Granger Causal Attribution}
Granger causality~\cite{granger1969,granger1980testing} is a concept of causality based on prediction, which declares that if a time series $\vx$ Granger-causes a time series $\vy$, then $\vy$ can be better predicted using all available information than if the information apart from $\vx$ had been used. 

\begin{figure}[t]
\centering
\includegraphics[width=0.6\linewidth]{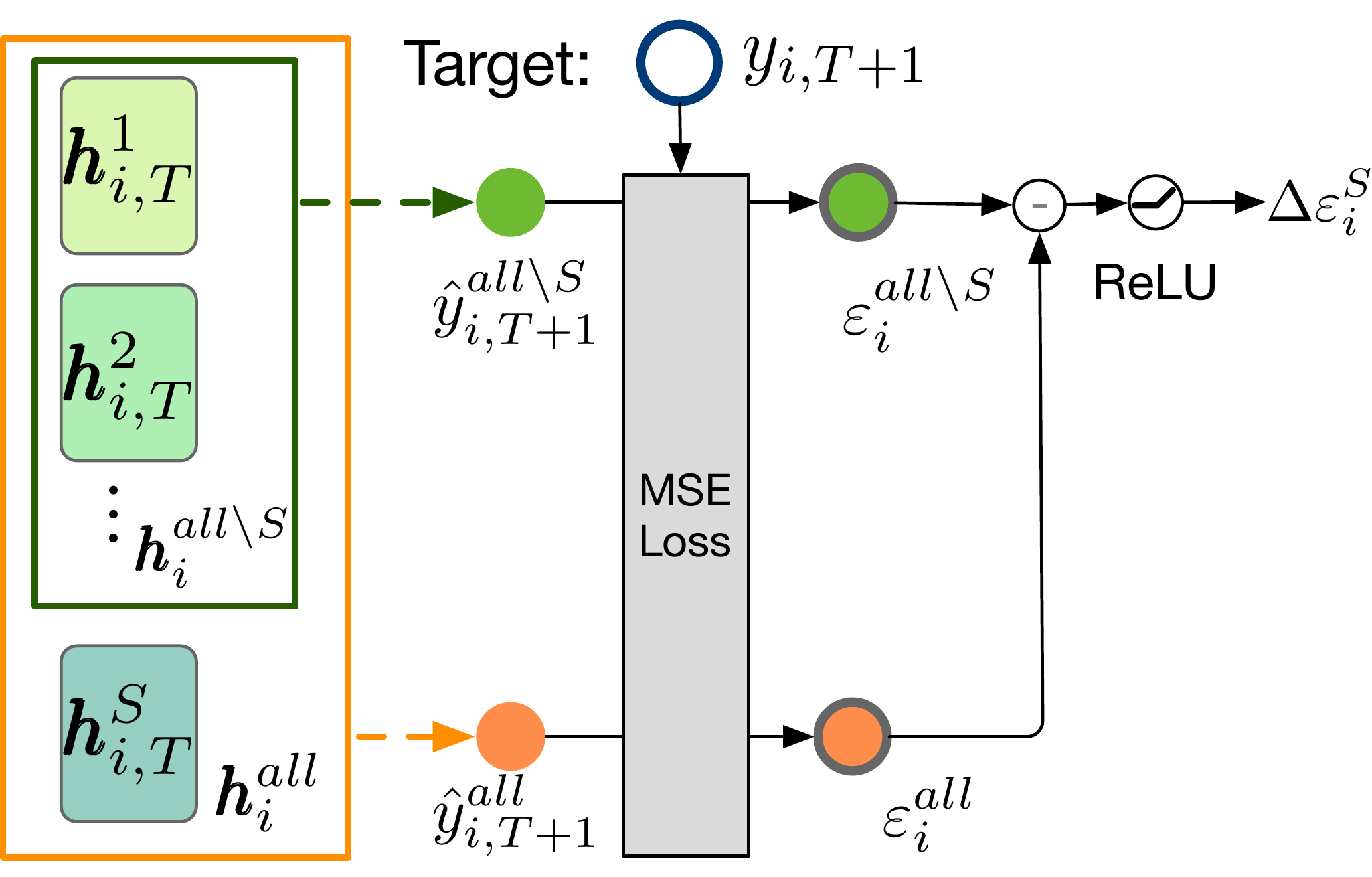} 
\caption{The Granger causal attention mechanism.}
\label{fig:gc}  
\end{figure}

As illustrated in Figure~\ref{fig:gc}, we obtain the Granger causal attributions by comparing the prediction error when using all available information with the error when using the information excluding one variable series. In particular, given all the hidden embeddings  $\{\vh_{i,T}^s\}_{s=1}^S$ of the $i$-th individual, we obtain the embedding that encodes all available information and the one that encodes all available information excluding one variable $s$, denoted as $\vh^{all}_i$ and $\vh_i^{all\backslash{s}}$ respectively, by concatenating the embeddings of corresponding variables:
\begin{align}
    &\vh^{all}_i=[\vh_{i,T}^j]_{j=1}^S,\quad
    \vh_i^{all\backslash{s}}=[\vh_{i,T}^j]_{j=1,j\not=s}^S,
\end{align}
where $[\cdot]$ represents the concatenation operation.
Then we feed them into respective predictors, denoted as $g_{all}(\cdot)$ and $g_s(\cdot)$, to get the predicted value of target and compute the squared errors: 
\begin{align}
    \hat{y}_{i,T+1}^{all}=g_{all}(\vh_i^{all}),&\quad\hat{y}_{i,T+1}^{all\backslash{s}}=g_{s}(\vh_i^{all\backslash{s}}),\\
    \varepsilon_i^{all}=(\hat{y}_{i,T+1}^{all}-y_{i,T+1})^2,&\quad\varepsilon_i^{all\backslash{s}}=(\hat{y}_{i,T+1}^{all\backslash{s}}-y_{i,T+1})^2,
\end{align}
where the predictor $g_{all}(\cdot)$ and $g_s(\cdot)$ can be MLP networks. Inspired by~\cite{schwab2019granger}, we define the Granger causal attribution of the target variable corresponding to the $s$-th variable as the decrease in error when adding the $s$-th series to the set of available information, computed as:
\begin{equation}\label{eq:granger_dec}
\Delta\varepsilon_{i}^s=ReLU(\varepsilon^{{all}\backslash{s}}_i-\varepsilon_{i}^{all}),
\end{equation}
where $ReLU(\cdot)$ is the rectified linear unit. \revise{Since the error decrease is supposed to be a non-negative value, we use $ReLU(\cdot)$ to constrain the non-negativity during the training phase}. For each individual $i$, by normalising the Granger causal attribution, we obtain an attention vector, denoted as $\pmb{q}_i$. The attention factor for the $s$-th variable can be computed as: 
\begin{equation}\label{eq:granger_att}
    q_i^s=\frac{\Delta\varepsilon_{i}^s}{\sum_{j=1}^S\Delta\varepsilon_{i}^j}.
\end{equation}
We can conclude from Equation~(\ref{eq:granger_dec}) and (\ref{eq:granger_att}) that if adding the $s$-th series to the existing time series does not improve the accuracy, then the attention factor $q_i^s$ is zero and the $s$-th time series is Granger noncausal for the target series. The attention factors in Equation (\ref{eq:granger_att}) can capture Granger causality, thus, we refer to the attention mechanism as Granger causal attention. 

\subsubsection{Prototype Learning}

\begin{figure}[t]
\centering 
\includegraphics[width=0.8\linewidth]{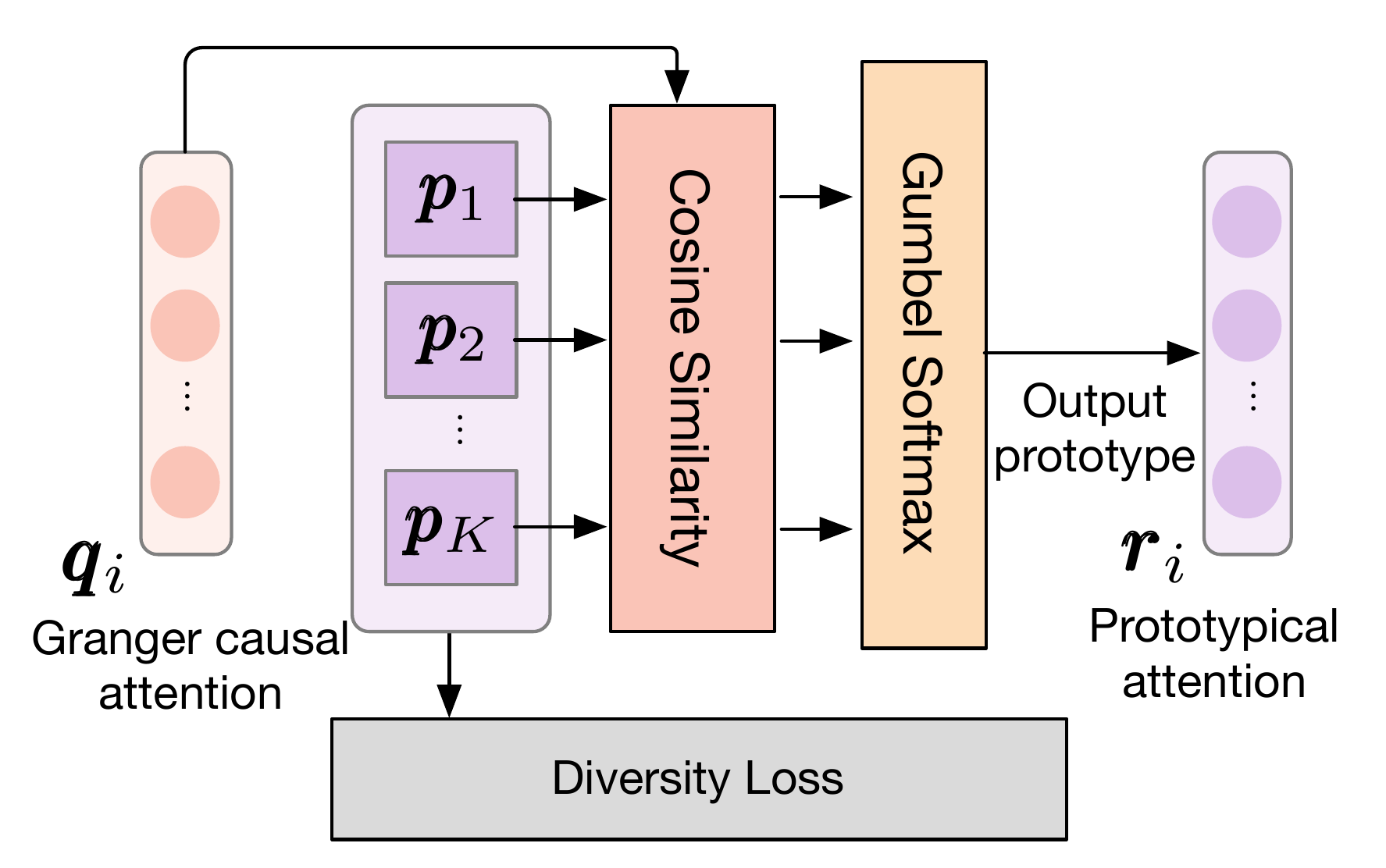} 
\caption{The Prototypical Granger Causal Attention.}
\label{fig:proto} 
\vspace{-5pt}
\end{figure}
The Granger causal attention above is not robust enough to reconstruct Granger causal structure, given limited data (e.g., very short time series) of each individual in training. We address the problem by generating Granger causal prototypes from all the individuals, under the assumption that there should be several common causal structures among heterogeneous individuals. The overview of the prototypical attention is illustrated in Figure~\ref{fig:proto}

\revise{In particular, we assume there exist $K$ common Granger causal structures, namely Granger causal prototypes, denoted as $\{\vp_k\}_{k=1}^K$. The prototype $\vp_k\in\mathbb{R}^S$ is a parameter vector to be optimized 
. The goal is to output one prototype most similar to the Granger causal attention. Thus, we first compute the similarity between the Granger causal attention vector $\pmb{q}_i$ of individual $i$ and each prototype vectors $\vp_k$:} 
\begin{equation}
    d_{k,i}=\frac{\vp_k\cdot\pmb{q}_i}{\|\vp_k\|\|\pmb{q}_i\|},
\end{equation}
Then we output a prototype most similar to $\vq_i$ by sampling from the similarity distribution $\vd_i$ using Gumbel-Softmax~\cite{maddison2016concrete,jang2016categorical}
:
\begin{equation}
    \ve=\text{GumbelSoftmax}(\vd_i)=\text{softmax}((\log(\vd_i)+\vg)/\tau),
\end{equation}
where $\text{GumbelSoftmax}(\cdot)$ denotes the Gumbel-Softmax function, $\ve\in\mathbb{R}^K$ is the sample vector which approaches one-hot, and $\vg$ is a vector of i.i.d. samples drawn from Gumbel$(0, 1)$ distribution. The parameter $\tau$ is the softmax temperature, and the distribution becomes discrete when $\tau$ goes to 0.
With the sample vector $\ve$, the output prototype $\hat{\vp}$ can be obtained as $\hat{\vp} = [\vp_1,\vp_2,\dots,\vp_K]\cdot\ve.$
After normalizing the sampled prototype, we obtain an attention vector for individual $i$, denoted as $\vr_i$, namely prototypical attention.

The Granger causal attention reflects the Granger causal structure specific to each individual, while the prototypical attention reflects one common Granger causal structure most similar to the Granger causal structure of each individual. To detect the specific and common causal structures simultaneously, we summarize them together and generate the prototypical Granger causal attention $\va_i$ as follows: 
\begin{equation}
    \va_i = \alpha\pmb{q}_i+(1-\alpha)\pmb{r}_i,
\label{eq:att}
\end{equation}
where $\alpha\in[0,1]$ is a hyperparameter that controls the ratio of the two attention mechanism.

Finally, the prediction of the target variable's next value can be computed as the weighted sum of the predicted values from all variables:
\begin{equation}\label{eq:pred_y}
    \hat{y}_{i,T+1}=\sum_{s=1}^S{a_i^s}\hat{y}_{i,T+1}^s.
\end{equation}
In practice, we notice that if the data is sparse and the prototype learning module can help enhance the information sharing. However, if the data is abundant then the incremental improvements brought by this module are marginal and we can discard it for both simplicity and efficiency. 

\vspace{-2pt}
\subsection{The Algorithm}
In order to obtain accurate prediction and Granger causality structure, and generate diverse common causality structures, the objectives of InGRA consist of three parts. The first two objective functions are to encourage accurate predictors, including the predictors $f(\cdot)$ to perform final prediction and the auxiliary predictors $g(\cdot)$ to compute Granger attribution, and we adopt the 
the mean squared error (MSE) as the prediction loss function:
\begin{equation}\label{eq:pre_loss}
     \mathcal{L}_\text{pred}=\frac{1}{N}\sum_{i=1}^N(\hat{y}_{i,T+1}-y_{i,T+1})^2,
     \mathcal{L}_\text{aux}=\frac{1}{N}\sum_{i=1}^N(\varepsilon_i^{all}+\sum_{s=1}^s\varepsilon_i^{all\backslash{s}}).
\end{equation}
The last objective function is to avoid duplicate prototypes by a diversity regularization term that penalizes on prototypes that are similar to each other~\cite{ming2019interpretable}:
\begin{equation}
    \mathcal{L}_\text{div}=\sum_{i=1}^K\sum_{j=i+1}^K\max(\gamma,\frac{\vp_i\cdot\pmb{p}_j}{\|\vp_i\|\|\pmb{p}_j\|}),
\end{equation}
where $\gamma$ controls the closeness to a tolerable degree.

To summarize, the loss function, denoted by
$\mathcal{L}$, is given by:
\begin{equation}
    \mathcal{L}=\mathcal{L}_\text{pred}+\lambda_1\mathcal{L}_\text{aux}+\lambda_2\mathcal{L}_\text{div},
\end{equation}
where $\lambda_1$ and $\lambda_2$ are hyperparameters that adjust the ratios between the losses.

We adopt stochastic gradient descent (SGD) to optimize the network parameters and the prototype parameters. To initialize the prototypes, we first pretrain InGRA for several epochs, and then employ $k$-means with cosine similarity on the Granger causal attentions $\{\vq\}_{i=1}^N$, and finally we take the cluster centers as the initial prototypes.  
\revise{In the implementation, given a long MTS of length $L$, we first slide a window of length $T+1$ over the input MTS to generate series data $\{\vx^s_{i,t:t+T}\}_{t=1}^{L-T}$, and then split the dataset into training and testing sets. We train our model on the training set, and obtain the network parameters and prototype parameters. By feeding the testing set into the trained InGRA model, we obtain the Granger causal attention $\vq_i$ by Equation ~(\ref{eq:granger_att}) and the predicted value $\hat{y}_{i,T+1}$ according to Equation ~(\ref{eq:pred_y}).} 


%% file: 4.experiment.tex
In this section, we evaluate the Granger causal structure reconstruction performance on multivariate time series from both single individual and multiple individuals, which we refer to as homogeneous and heterogeneous MTS respectively. For heterogeneous MTS, we also visualize the learned structures, prototypes, and the Granger causal attention vectors to give an intuitive understanding of how InGRA works and how the Granger causal structure is being built. Moreover, we conduct an online A/B test on an E-commerce advertising platform to further test InGRA in more practical situations. 
\subsection{Experimental Setup} 
We first evaluate the Granger causal structure recovery performances on homogeneous MTS with two causal benchmark datasets.
 
\textbf{Finance}\footnote{http://www.skleinberg.org/data.html} \cite{kleinberg2009causality} consists of simulated financial market time series with known underlying causal structures. Each dataset includes 25 variables of length 4,000. For each dataset, we choose variables that are related to the most causes as the target variables to test model abilities in the relatively most challenging scenarios. 

\textbf{FMRI}\footnote{We use the processed FMRI data provide by ~\cite{nauta2019tcdf}.}~\cite{smith2011network} contains 28 different Blood-oxygen-level dependent time series datasets with the ground-truth causal structures. We use the first 5 datasets and take the first variable as the target as causal variables distribute relatively evenly in this dataset. 

We evaluate the performance on heterogeneous MTS with the following synthetic data.

\textbf{Synthetic data}: We first obtain the $S$ time series through the following Non-linear Autoregressive Moving Average (NARMA)~\cite{atiya2000new} generators:
\begin{equation}\small
    x_{i,t}^s = {\alpha_s}x_{i,t-1}^s + {\beta_s}x_{i,t-1}^s\sum_{j=1}^{d}x_{i,t-j}^s + {\gamma_s}\varepsilon_{i,t-d}\varepsilon_{i,t-1} + \varepsilon_{i,t},
\label{eq:syn1}
\end{equation}
where $\varepsilon_{i,t}$ are zero-mean noise terms of 0.01 variance, $d$ is the order of non-linear interactions, and $\alpha_s$, $\beta_s$ and $\gamma_s$ are parameters specific to variable $s$, generated from $\mathcal{N}(0,0.1)$. Then, we generate the target series from the generated exogenous series via the formula:
\begin{equation}\small
    y_{i,t}=\sum_{s=1}^S{\omega_i^s(\pmb{\eta}_i^s})^\mathsf{T}\tanh{(\vx_{i,t-p:t-1}^s)}+\varepsilon_{i,t},
    \label{eq:syn2}
\end{equation}
where $\omega_i^s\in\{0,1\}$ controls the underlying causal relationship from the $s$-th variable to the target variable, $\pmb{\eta}_i^s\in\mathbb{R}^p$ controls the causal strength sampling from $\mbox{Unif}\{-1, 1\}$, and $\vx_{i,t-p:t-1}^s\in\mathbb{R}^p$ represents the last $p$ historical values of variable $s$ of sample $i$. The 0-1 indicator vector $\pmb{\omega}_i=(\omega_i^1,\omega_i^2,\dots,\omega_i^S)^\mathsf{T}\in\mathbb{R}^S$ is the ground-truth causal structure of $i$-th individual.




We compare our method with previous causal discovery methods including linear Granger causality~\cite{granger1969,lutkepohl2005new} and TCDF~\cite{nauta2019tcdf}, as well as the interpretable neural network based prediction method, i.e., IMV-LSTM~\cite{guo2019exploring}, using the standard metrics of Average Precision (AP), and Area Under the ROC Curve (ROC-AUC)~\cite{fawcett2006introduction}. 
\begin{itemize}[leftmargin=*]
\item\textbf{Linear Granger}~\cite{granger1969,granger1980testing}: We conduct a Granger causality test in the context of Vector Autoregression (VAR) as described in chapter 7.6.3 in ~\cite{lutkepohl2005new} and implemented by the Statsmodels package~\cite{seabold2010statsmodels}. In detail, we perform F-test at 5\% significance level. The maximum number of lags to check for order selection is set to 5, which is larger than the causal order in the ground-truth.

\item\textbf{TCDF}~\cite{nauta2019tcdf}: TCDF\footnote{\url{https://github.com/M-Nauta/TCDF}} learns causal structure on multivariate time series by attention-based convolutional neural networks combined with a causal validation step. We follow the default settings as described in~\cite{nauta2019tcdf}
.


\item\textbf{IMV-LSTM}~\cite{guo2019exploring}:  IMV-LSTM\footnote{\url{https://github.com/KurochkinAlexey/IMV_LSTM}} is a multi-variable attention-based LSTM model capable of both prediction and variable importance interpretation, with the attention factors reflecting importance of variables in prediction. Thus, we take the learnt attention vectors as the Granger causal weights in the experiment. IMV-LSTM is implemented by Adam optimizer with the mini-batch size 64, hidden layer size 128 and learning rate 0.001.
\end{itemize}

\subsection{
Performance on Homogeneous MTS}
To evaluate the Granger causal discovery performance on homogeneous multivariate time series, we train individual models for each dataset with the hyper-parameter $\alpha$ equaling  0.5. We report AP and ROC-AUC averaged across all datasets, with the standard deviation reported in Table~\ref{tab:single}. As can be seen, the proposed method greatly surpasses other methods. Especially, InGRA recovers the ground-truth causal structure with high scores on the Finance data.

\begin{table}[h]
\centering
\setlength\abovecaptionskip{0pt}
\caption{Granger causal structure reconstruction results on Finance and FMRI data.} 
\label{tab:single}
\resizebox{\linewidth}{!}{
\begin{tabular}{@{}ccccc@{}}
\toprule
\multirow{2}{*}{Methods} & \multicolumn{2}{c}{Finance (9 datasets)}                                 & \multicolumn{2}{c}{FMRI (5 datasets)}                                \\ \cmidrule(l){2-5} 
                         & AP                       & ROC-AUC                      & AP                       & ROC-AUC                      \\ \midrule
IMV-LSTM                 & 0.778$\pm$0.222 & 0.862$\pm$0.172 & 0.593$\pm$0.239 & 0.620$\pm$0.136 \\
linear Granger           & 0.187$\pm$0.036 & 0.652$\pm$0.084 & 0.492$\pm$0.310  & 0.654$\pm$0.126 \\
TCDF                     & 0.478$\pm$0.263 & 0.766$\pm$0.145 & 0.540$\pm$0.250 & 0.664$\pm$0.099 \\
InGRA                    & \textbf{1.000$\pm$0.000}$^{**}$ & \textbf{1.000$\pm$0.000}$^{*}$ & \textbf{0.641$\pm$0.327} & \textbf{0.740$\pm$0.122} \\ \bottomrule
\end{tabular}}\\
\footnotesize{$^{**}$ and $^{*}$ denotes the p-value is less than 1\% and 5\% respectively.}
\setlength\belowcaptionskip{-5pt}
\end{table}

\vspace{-5pt}
\subsection{
Performance on Heterogeneous MTS}
\revise{We evaluate the Granger causal discovery performance and the inductive capacity on heterogeneous multivariate time series. We denote the number of common causal structures as $C$, the number of variables as $S$ and the series length as $T$, and generate $100$ multivariate time series for each common causal structure according to Equation (\ref{eq:syn1}) and Equation (\ref{eq:syn2}), forming 100$C$ datasets, with 20\% datasets as unseen series and others as training series. For the inductive methods InGRA and IMV-LSTM, we train one model using all the training series, while for other methods, we train separate models for each dataset.} 
\begin{table*}[]
\centering\setlength\abovecaptionskip{0pt}
\caption{Granger causal structure reconstruction results w.r.t the variable number ($C$=3, $T$=1000).}
\label{tab:S}
\resizebox{0.9\textwidth}{!}{\begin{tabular}{@{}ccccccccc@{}}
\toprule
\multirow{2}{*}{Methods} & \multicolumn{2}{c}{$S$=5} & \multicolumn{2}{c}{$S$=10} & \multicolumn{2}{c}{$S$=20}  \\ \cmidrule(l){2-7} 
                             & AP     & ROC-AUC     & AP     & ROC-AUC     & AP     & ROC-AUC     \\ \midrule
IMV-LSTM                          &    0.511$\pm$0.102        &  0.500$\pm$0.236           &   0.536$\pm$0.056        &      0.514$\pm$0.019      &     0.599$\pm$0.087       &   0.619$\pm$0.087       \\
linear Granger                     &    0.666$\pm$0.107   &    0.822$\pm$0.075    &     0.765$\pm$0.109       &    0.889$\pm$0.063         &    0.826$\pm$0.106        &        0.854$\pm$0.080     \\
TCDF                              &   0.523$\pm$0.103         &    0.523$\pm$0.214      &     0.548$\pm$0.165     &  0.587$\pm$0.180           &    0.584$\pm$0.162       &     0.642$\pm$0.152        \\ \midrule
InGRA ($\alpha=1$)       	            &   0.886$\pm$0.177$^{**}$    &  0.906$\pm$0.143$^{**}$      &   	0.974$\pm$0.038$^{**}$	         &    0.975$\pm$0.037$^{**}$	         &   0.830$\pm$0.108	         &     0.883$\pm$0.069$^{**}$        \\
InGRA ($\alpha=0.5$)                              &   \textbf{0.911$\pm$0.147}$^{**}$	    &  \textbf{0.922$\pm$0.122}$^{**}$      &    	\textbf{0.998$\pm$0.009}$^{**}$       &  	\textbf{0.999$\pm$0.008}$^{**}$	         &       \textbf{0.858$\pm$0.103}$^{*}$     &        	\textbf{0.939$\pm$0.050}$^{**}$        \\\bottomrule
\end{tabular}}
\\
{\footnotesize{$^{**}$ denotes the p-value is less than 1\%, and $^{*}$ denotes the p-value is less than 5\%.}}
\centering
\end{table*}

\begin{table*}[]
\centering
\setlength\abovecaptionskip{0pt}\setlength{\abovecaptionskip}{0pt} 
\setlength{\belowcaptionskip}{0pt} 
\caption{Granger causal structure reconstruction results w.r.t the series length  ($C$=3, $S$=10).}
\label{tab:T}
\resizebox{0.9\textwidth}{!}{\begin{tabular}{@{}ccccccccc@{}}
\toprule
\multirow{2}{*}{Methods} & \multicolumn{2}{c}{$T$=20} & \multicolumn{2}{c}{$T$=100} & \multicolumn{2}{c}{$T$=1000}  \\ \cmidrule(l){2-7} 
                             & AP     & ROC-AUC     & AP     & ROC-AUC     & AP     & ROC-AUC     \\ \midrule
IMV-LSTM                          &     0.467$\pm$0.025     &    0.541$\pm$0.035        &  0.503$\pm$0.081        &    0.511$\pm$0.018       &            0.536$\pm$0.056        &      0.514$\pm$0.019            \\
linear Granger                     &  0.400$\pm$0.000     &   0.500$\pm$0.000    &           0.889$\pm$0.152     &   0.943$\pm$0.085            &          0.765$\pm$0.109       &    0.889$\pm$0.063             \\
TCDF                              &    0.518$\pm$0.131       &      0.513$\pm$0.112     &         0.517$\pm$0.120   &     0.544$\pm$0.166        &           0.548$\pm$0.165     &  0.587$\pm$0.180            \\ \midrule
InGRA ($\alpha=1$)       	            & 0.790$\pm$0.142$^{**}$ & 0.793$\pm$0.150$^{**}$  &   		           0.973$\pm$0.038$^{**}$	         &   0.974$\pm$0.038$^{*}$	   &      0.974$\pm$0.038$^{**}$	         &    0.975$\pm$0.037$^{**}$            \\
InGRA ($\alpha=0.5$)       &  \textbf{0.824$\pm$0.123}$^{**}$      &   \textbf{0.833$\pm$0.117}$^{**}$      &    \textbf{0.973$\pm$0.040}$^{**}$	       &  \textbf{0.976$\pm$0.036}$^{**}$		&      	\textbf{0.998$\pm$0.009}$^{**}$       &  \textbf{0.999$\pm$0.008}$^{**}$                  	        \\\bottomrule
\end{tabular}}\\
{\footnotesize{$^{**}$ denotes the p-value is less than 1\%, and $^{*}$ denotes the p-value is less than 5\%.}}
\end{table*}

\begin{table*}[t!]
\centering\setlength\abovecaptionskip{0pt}
\caption{Granger causal structure reconstruction results w.r.t. the common structure number $C$ ($S$=10, $T$=1000). We set the hyper-parameter of prototype number $K$ in the model as the same as the ground-truth common structure number $C$.}
\label{tab:K}
\resizebox{0.8\textwidth}{!}{
\begin{tabular}{@{}ccccccc@{}}
\toprule
\multirow{2}{*}{Methods}                                      & \multicolumn{2}{c}{$C$=3}                                     & \multicolumn{2}{c}{$C$=5}                                     & \multicolumn{2}{c}{$C$=7}                                    \\ \cmidrule(l){2-7} 
                                               & AP                       & ROC-AUC                      & AP                       & ROC-AUC                      & AP                       & ROC-AUC                      \\ \midrule
InGRA($\alpha=1$)                   & 0.974$\pm$0.038 & 0.975$\pm$0.037 & 0.891$\pm$0.118 & 0.883$\pm$0.128 & 0.838$\pm$0.118 & 0.850$\pm$0.113 \\
InGRA($\alpha=0.5$)                   & 0.998$\pm$0.009 & 0.999$\pm$0.008 & 0.924$\pm$0.091 & 0.913$\pm$0.105 &  0.851$\pm$0.113  & 0.855$\pm$0.099 \\ \bottomrule
\end{tabular}}
\end{table*}

We report AP and ROC-AUC results w.r.t the variable number, the series length and the common structure number in Table~\ref{tab:S} to~\ref{tab:K}, respectively. We observe that InGRA outperforms other methods significantly in all cases, and InGRA ($\alpha=0.5$) (with the Prototypical Granger causal attention) performs better than InGRA ($\alpha=1$) (only with Granger causal attention). The observations demonstrate the superior causal discovery performance of InGRA, the effectiveness of the prototypical Granger causal attention in InGRA, and the advantages of utilizing shared commonalities among heterogeneous MTS. Regarding the other competitors, linear Granger performs the best followed by TCDF and IMV-LSTM at most cases. The possible reason is that linear Granger can detect Granger causal relations to some extent, though it utilizes linear model, i.e., Vector autoregression (VAR). TCDF utilizes attention-based CNN to inference potential causals followed by a causal validation step, but the attention it proposed cannot reflect Granger causality, thus achieves unsatisfactory performance. Compare to the performance on homogeneous multivariate time series, the performance of IMV-LSTM drops dramatically, which indicates that the attention mechanism in IMV-LSTM fails given heterogeneous multivariate time series.

In Table~\ref{tab:S}, we vary the number of variables to generate datasets of different complexity, and we can see that InGRA outperforms other competitors consistently across different $S$, and achieves good performance when $S$ is as large as 20, demonstrating our method's capability to infer complex causal structures. Since in practice, the size of collected data is often limited, which poses challenges to recover causal structure, thus we also vary the length of time series to see the model robustness to data of small sizes. As can be seen in Table~\ref{tab:T}, InGRA outperforms other methods across all cases, even when $T$ is as small as 20, which demonstrates that advantage of using shared information. We also observe that InGRA ($\alpha=0.5$) surpasses InGRA ($\alpha=1$) by a large margin, which demonstrates the learning prototypical attention can alleviate the over-fitting problem. In Table~\ref{tab:K}, we control the causal heterogeneity by varying the number of common causal structures $C=\{3,5,7\}$. We observe that the performance of InGRA decreases with increasing $C$.

\begin{figure}[]
\centering\setlength{\abovecaptionskip}{0pt}
\setlength{\belowcaptionskip}{-5pt} 
\includegraphics[width=0.5\linewidth]{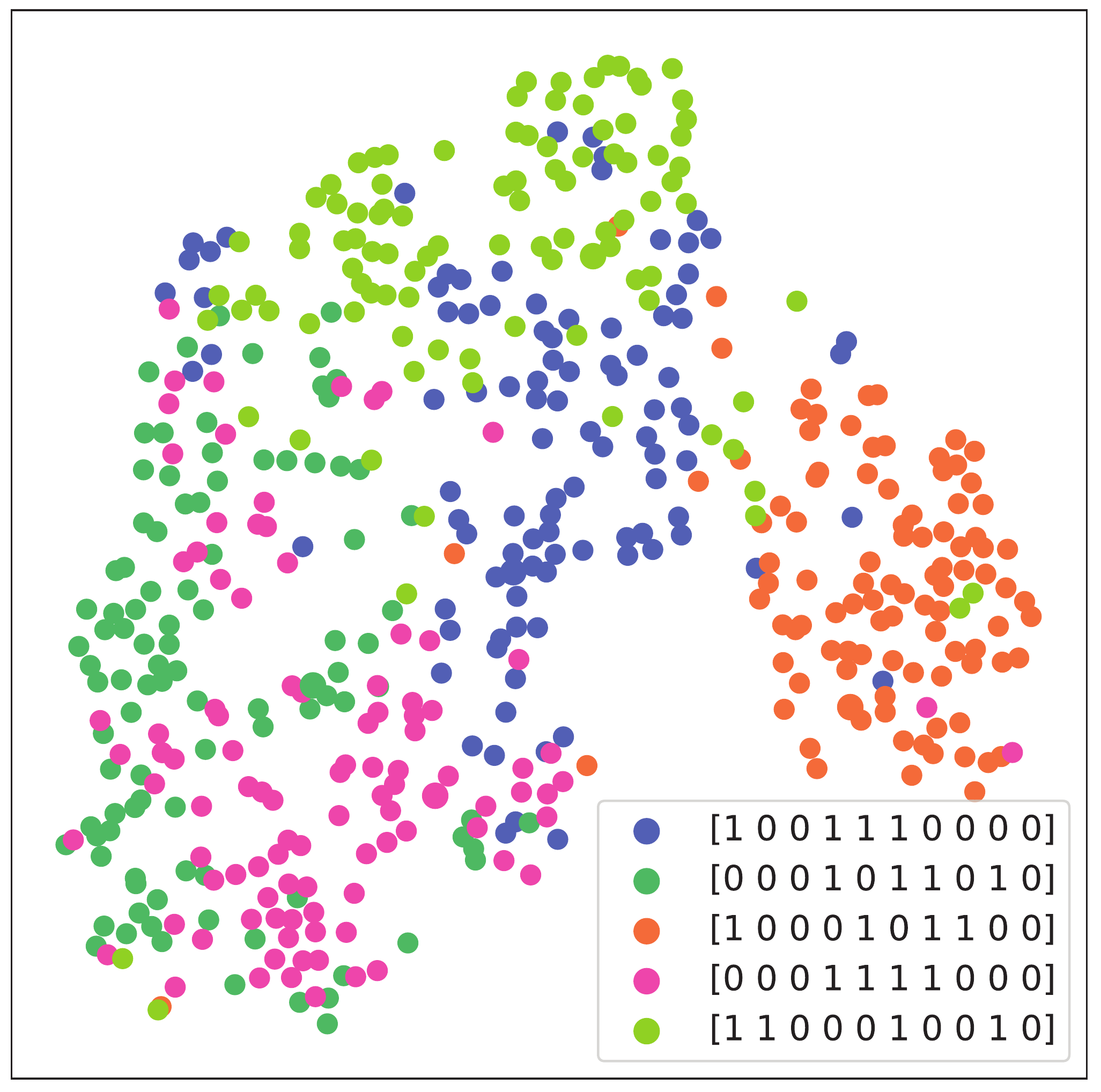} 
\caption{2D t-SNE projections of attention vectors from 500 individuals. Color of a node indicates the underlying causal structures. The causal groundtruth $\pmb{\omega}_i$ is shown in the legend ($S$=10,$T$=100).} 
\label{fig:tsne}
\end{figure}
\vspace{-5pt}
\begin{figure}[]
\centering\setlength{\abovecaptionskip}{-5pt} 
\setlength{\belowcaptionskip}{-10pt} 
\centering
\includegraphics[width=0.7\linewidth]{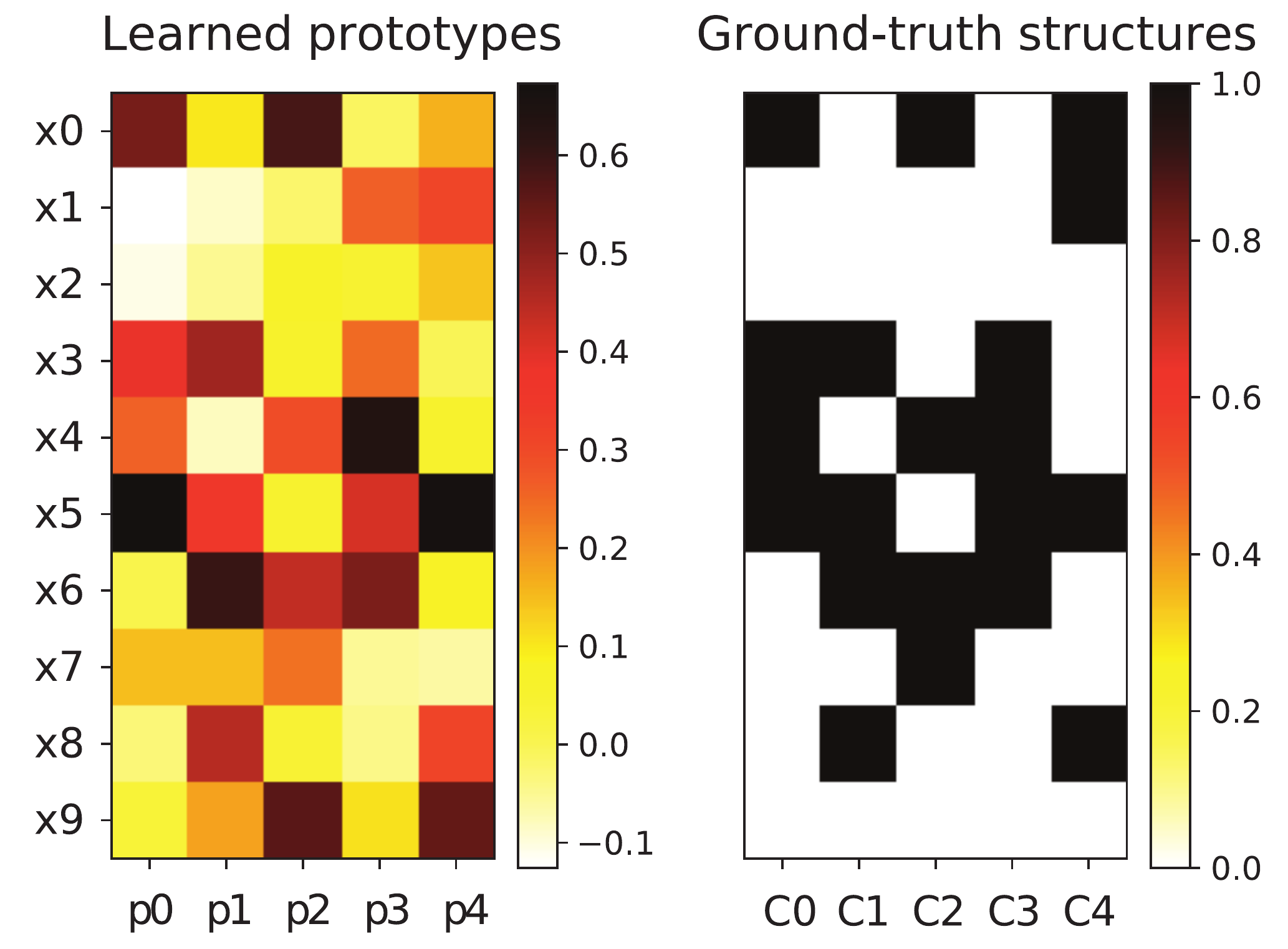}
\caption{The visualization of prototypes and ground-truth causal structures. Plots are shown for various number of common causal structures $C$ ($S$=10, $T$=1000). Each column of the heat map visualizes one structure.}
\label{fig:vis_proto}
\vspace{-10pt}
\end{figure}

\subsection{Visualization}
\subsubsection{Visualization of Learned Structures}
In Figure~\ref{fig:tsne}, we map the learned causal attention vectors to a 2D space by the visualization tool t-SNE~\cite{maaten2008visualizing}. Individuals of different causal structures are labeled by different colors. From the results, we observe that nodes belonging to the same causal structures are clustered together, which also demonstrates the effectiveness of our method. 


\subsubsection{Visualization of Learned Prototypes}
We visualize the the learned prototypes and the ground-truth causal structures in Figure~\ref{fig:vis_proto}. In this experiments, we set the hyper-parameter of  prototype number $K$ equal to the ground-truth common structure number $C$. From the results, we can see that the learned prototypes are similar to the ground-truth causal structures, which demonstrates the learned prototypes are interpretable.

\vspace{-1pt}

\subsection{Online A/B Tests}
In order to further evaluate the effectiveness of InGRA in practice, an online A/B test is conducted on an E-commercial platform, and the process is designed as follows:

We first train InGRA on the historical MTS of 30,665 items. Each MTS includes $26$ variables related to searching, recommending and advertising, such as Page View (PV), Gross Merchandise Volume (GMV) and Impression Position In-Page, etc. Here, we take the item popularity as the target series, and generate the underlying causal structure for each item. We sample 100 items whose impression position in-page Granger-causes the item popularity with high confidence, and divide them into two buckets randomly. For Bucket A, we adjust the impression positions in-page of each item by one grid since 2019/08/19 till 2019/08/29, and ensure the intervention has little impact on other variables. For Bucket B, we do nothing. 

\begin{figure}[]
\centering\setlength{\abovecaptionskip}{0pt} 
\includegraphics[width=0.9\linewidth]{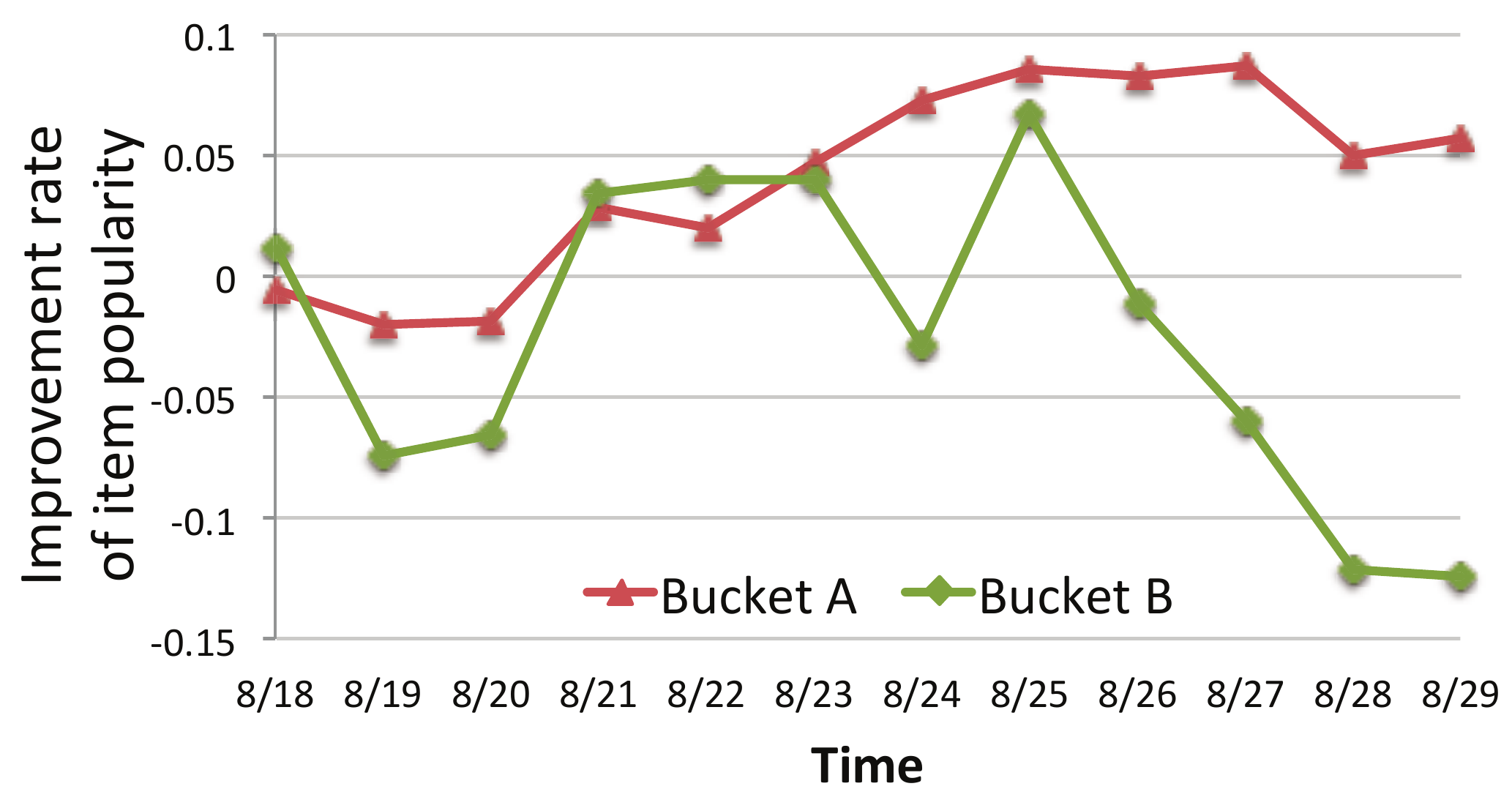} 
\caption{The result of online A/B test. The intervention starts from 08/19, and results in the item popularity improvement rate of Bucket A consistently outperforming Bucket B after 8/22, and the gap between the two buckets increases significantly since 08/25. } 
\setlength{\belowcaptionskip}{-10pt}
\vspace{-4pt}
\label{fig:online}
\end{figure}

As shown in Figure~\ref{fig:online}, four days after the beginning of the intervention, the item popularity improvement rate of Bucket A consistently outperforms that of Bucket B, and the gap between the two buckets increases significantly since 2019/08/25, which shows that the intervention, i.e., adjusting the impression positions in-page, causes the improvement on item popularities, thus demonstrates the right causal relationships detected by InGRA.

%% file: 5.conclusion.tex
We formalize the problem of inductive Granger causal modeling on multivariate time series and propose an inductive framework InGRA to solve it. In particular, we propose a novel attention mechanism, namely prototypical Granger causal attention, which computes Granger causal attribution combined with prototype learning, to reconstruct Granger causal structures and uncover common causal structures.
The approach has been successfully evaluated by offline experiments on benchmark datasets compared to previous methods, also confirmed by an online A/B test on an E-commercial platform. 
A particularly interesting direction for future work is exploring the time delay between a Granger cause and the occurrence of its effect. 